\pdfoutput=1
\documentclass[10pt,twocolumn,letterpaper]{article}

\usepackage[final,pagenumbers]{cvpr} 

\usepackage{graphicx}
\usepackage{comment}
\usepackage{epsfig}
\usepackage{color}
\usepackage{booktabs} 
\usepackage{amsmath,epsfig,bm,amssymb}
\usepackage{algpseudocode}
\usepackage{multirow}
\usepackage{longtable}
\usepackage{makecell}
\usepackage{subfloat}
\usepackage{cases}
\usepackage{diagbox}
\usepackage{mathrsfs}
\usepackage{multirow}
\usepackage{caption}
\usepackage{cite}
\usepackage{cases}
\usepackage{diagbox}
\usepackage{algorithm}
\usepackage{amssymb}

\usepackage{pifont}
\newcommand{\cmark}{\ding{51}}
\newcommand{\xmark}{\ding{55}}

\def\eqref#1{equation~\ref{#1}}

\def\1{\bm{1}}

\def\rvd{{\mathbf{d}}}

\def\rvp{{\mathbf{p}}}

\def\rvt{{\mathbf{t}}}

\def\rvv{{\mathbf{v}}}

\def\rvx{{\mathbf{x}}}

\def\rvz{{\mathbf{z}}}

\DeclareMathAlphabet{\mathsfit}{\encodingdefault}{\sfdefault}{m}{sl}
\SetMathAlphabet{\mathsfit}{bold}{\encodingdefault}{\sfdefault}{bx}{n}

\def\gD{{\mathcal{D}}}

\def\gL{{\mathcal{L}}}

\usepackage[dvipsnames]{xcolor}

\makeatletter
\newcommand{\namelong}[1]{DISCO}
\newcommand{\nameshort}[1]{DISCO}
\newcommand{\namelongm}[1]{DISCO}
\newcommand{\nameshortm}[1]{DISCO}
\newcommand*{\affaddr}[1]{#1} 
\newcommand*{\affmark}[1][*]{\textsuperscript{#1}}
\newcommand{\printfnsymbol}[1]{%
  \textsuperscript{\@fnsymbol{#1}}%
}
\definecolor{da}{RGB}{220, 248, 224}
\definecolor{ds}{RGB}{187, 213, 255}
\definecolor{c}{RGB}{255, 204, 204}
\usepackage[pagebackref,breaklinks,colorlinks]{hyperref}

\usepackage[capitalize]{cleveref}
\crefname{section}{Sec.}{Secs.}
\Crefname{section}{Section}{Sections}
\Crefname{table}{Table}{Tables}
\crefname{table}{Tab.}{Tabs.}
\newlength\savewidth\newcommand\shline{\noalign{\global\savewidth\arrayrulewidth
  \global\arrayrulewidth 1pt}\hline\noalign{\global\arrayrulewidth\savewidth}}
\newcommand{\tablestyle}[2]{\setlength{\tabcolsep}{#1}\renewcommand{\arraystretch}{#2}\centering\footnotesize}

\def\cvprPaperID{4292} 
\def\confName{CVPR}
\def\confYear{2022}

\begin{document}

\title{Domain Adaptation via Prompt Learning}

\author{Chunjiang Ge\affmark[1] \quad Rui Huang\affmark[1] \quad Mixue Xie\affmark[2] \quad Zihang Lai\affmark[3] \\
Shiji Song\affmark[1] \quad Shuang Li \affmark[2] \quad Gao Huang \affmark[1]\textsuperscript{,}\affmark[4]\\
\affaddr{\small\affmark[1]Department of Automation, BNRist, Tsinghua University\quad}
\affaddr{\affmark[2]Beijing Institute of Technology\quad} \\
\affaddr{\small\affmark[3]Carnegie Mellon University\quad}
\affaddr{\affmark[4]Beijing Academy of Artificial Intelligence\quad}
}
\maketitle

\begin{abstract}
Unsupervised domain adaption (UDA) aims to adapt models learned from a well-annotated source domain to a target domain, where only unlabeled samples are given. Current UDA approaches learn domain-invariant features by aligning source and target feature spaces. Such alignments are imposed by constraints such as statistical discrepancy minimization or adversarial training. 
However, these constraints could lead to the distortion of semantic feature structures and loss of class discriminability. In this paper, we introduce a novel prompt learning paradigm for UDA, named Domain Adaptation via Prompt Learning~(DAPL). In contrast to prior works, our approach makes use of pre-trained vision-language models and optimizes only very few parameters. The main idea is to embed domain information into prompts, a form of representations generated from natural language, which is then used to perform classification. This domain information is shared only by images from the same domain, thereby dynamically adapting the classifier according to each domain. By adopting this paradigm, we show that our model not only outperforms previous methods on several cross-domain benchmarks but also is very efficient to train and easy to implement. 
\end{abstract}

 \section{Introduction}\label{sec:intro}

\begin{figure}[h]
    \centering
    \includegraphics[width=0.95\linewidth]{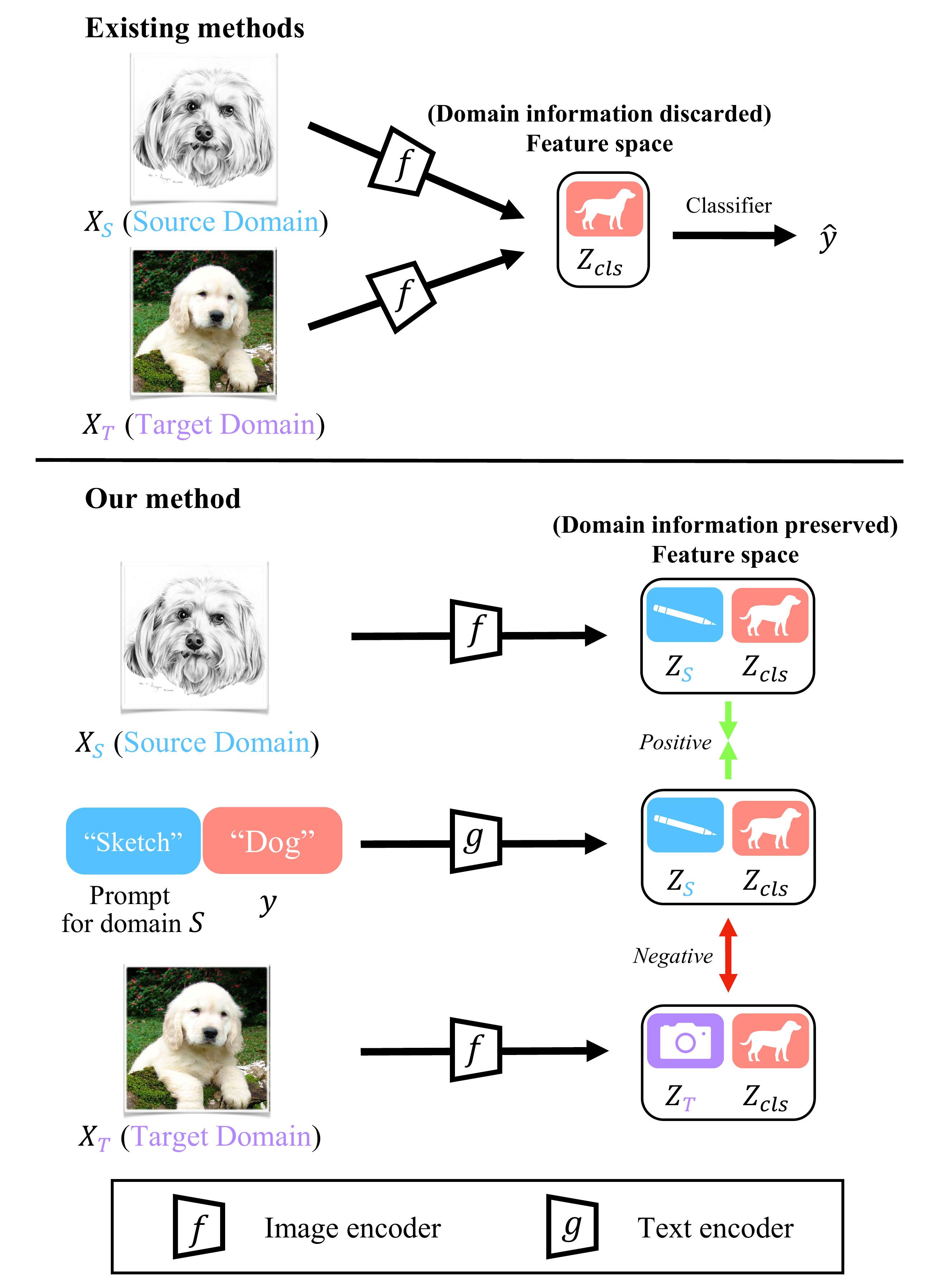}
    \caption{\textbf{Overview of DAPL.} We introduce the prompt tuning framework for domain adaptation. Top: conventional domain adaptation methods aim to remove domain-specific information via domain alignment or adversarial loss. This could lead to distorted feature representation when the manifold structures underlying the data distributions are complex~\cite{cai2019learning}. Bottom: Our method preserves domain information and tunes a prompt for each domain. Our model learns with a contrastive objective.}
    \label{fig:1}
    \vspace{-4mm}
\end{figure}

Deep Learning has achieved great success in recent years~\cite{he2016deep,huang2019convolutional} with the help of large-scale annotated datasets~\cite{deng2009imagenet}. Since annotating large-scale datasets is costly and time-consuming, researchers propose to train a model for an unlabeled domain by leveraging a related domain which is well-annotated. However, a model (\textit{e.g.}, a neural network) trained on an annotated domain may not generalize well to an unlabeled domain due to \textit{distribution shift}~\cite{DA_theory1, DA_theory2, DA_theory3}. The problem of Unsupervised Domain Adaptation (UDA) \cite{pan2009survey,ganin2015unsupervised,long2015learning} has been proposed to study the transferring of knowledge under such domain shift. 

Conventional UDA methods mainly resort to learning domain-invariant representations by aligning source and target domains. 
With similar features distribution led by domain alignment, the classifier trained on the source domain can be directly applied to the target data (\cref{fig:1}, top). One typical line of such methods is based on statistical discrepancy minimization \cite{tzeng2014deep,long2015learning, CMD_ICLR2017, long2017deep}, 
Maximum Mean Discrepancy (MMD)~\cite{long2015learning} and Central Moment Discrepancy (CMD) \cite{CMD_ICLR2017}.
Another typical line learns domain-invariant features via adversarial training by applying domain discriminators \cite{GAN_NIPS14,long2017conditional,JADA_MM19,lu2020stochastic}. Such methods confuse domain discriminators to reduce the difference between source and target domains in the feature space. However, reducing the discrepancy by aligning domains could lead to a loss of semantic information~\cite{tang2020unsupervised,DWL_CVPR21}. Such loss comes from
the entangled nature of semantic and domain information when the manifold structures of the data distributions are complex~\cite{cai2019learning}. To remedy this, some recent UDA methods \cite{li2018domain,BSP_ICML2019, SPL_AAAI20, tang2020unsupervised} advocate preserving the semantic information to maintain the class discriminability. 
However, these methods suffer from a subtle trade-off between \textit{domain alignment} and \textit{preserving semantic features}\cite{cai2019learning,stojanov2021domain,DWL_CVPR21} as two objectives could be adversarial. 
Learning disentangled semantic and domain representation could be an alternative since domain alignment could be discarded.

\begin{figure*}
    \centering
    \includegraphics[width=0.95\linewidth]{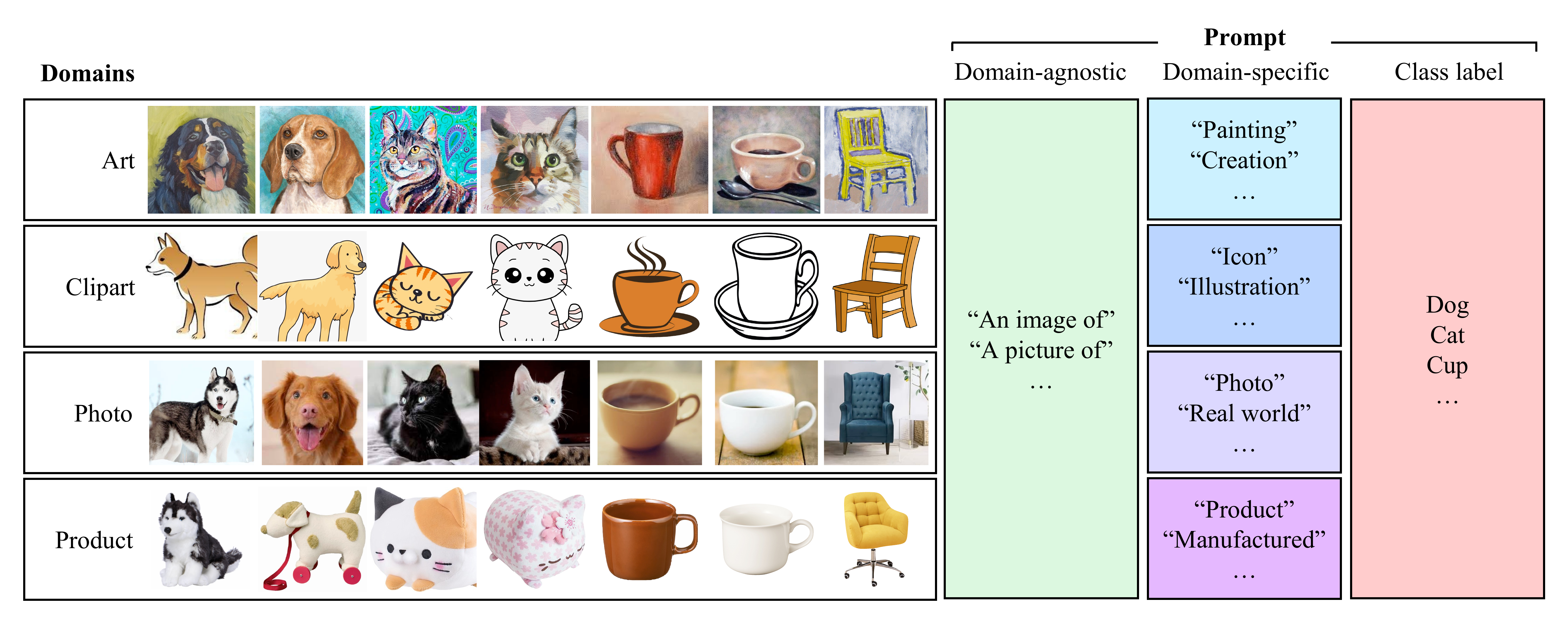}
    \caption{\textbf{Example prompt structure.} Our proposed prompt consists of three parts: (a) Domain-specific prompt; (b) Domain-agnostic prompt; (c) Class label. The first two parts are continuous and learned from data. The words shown here are for illustrative purposes.}
    \label{fig:2}
    \vspace{-4mm}
\end{figure*}

To learn disentangled \textit{semantic} and \textit{domain} representation, we introduce the prompt learning method\cite{liu2021gpt,liu2021pre,hu2021knowledgeable} to UDA, by learning a representation in a continuous label space. \cref{fig:2} illustrates our prompt design. The prompt consists of three parts: domain-agnostic context, domain-specific context, and class label~(token). Each image corresponds to a ground truth class through the class label of prompt. For example, an image that shows ``an art work of a dog'' could correspond to the prompt ``{\textcolor{LimeGreen}{An image of a}} {\textcolor{cyan}{painting}} {\textcolor{Melon}{Dog}}''. The domain-agnostic context represents general task information and is shared among all images. The domain-specific context represents domain information and is shared in each domain. The class label distinguishes different categories.

Such prompt learning method allows us to learn domain and category disentangled representation and avoids a loss of semantic information~\cite{tang2020unsupervised}.
We apply a contrastive objective for training~(\cref{fig:1}, bottom). An image and a text form a pair of positive examples only when the domain and category of them are matched respectively, while any other cases are negative examples. By contrasting the representation of $X_S$ and $y$, the image and text representation of the ``sketch" and ``dogs" are aligned in the feature space, respectively. Further, the text representation of ``sketch" is pushed away from the ``photo" domain by contrasting $X_T$ and $y$. More details are discussed in \cref{sec:disentangle}. Hence, the representation of domain and category are aligned respectively. We adopt \textit{Contrastive Language Image Pre-training}~(CLIP)~\cite{radford2021learning} as our backbone to facilitate prompt learning and contrastive learning. 


Extensive experiments on two classic cross-domain benchmarks demonstrate that our method consistently yields promising performance, \eg, we achieve an \textit{sota} performance of $74.5\%/86.9\%$ on Office-Home\cite{Office-Home} and VisDA-2017\cite{VisDA-2017}. To summarize, the contributions of our work are three-fold:


\begin{itemize}
    \item We propose Domain Adaptation via Prompt Learning~(\textbf{DAPL}) for unsupervised domain adaptation. To the best of our knowledge, we are the first to apply prompt learning in unsupervised domain adaptation.
    
    
    \item We propose to use domain-specific context in the prompt. Hence, we do not have to align domains at the cost of losing semantic information. Our method could learn continuous semantic representations for each category and domain.
    
    
    \item The proposed DAPL has achieved state-of-the-art performance on Office-Home and VisDA-2017 dataset, improving the accuracy by 
    $2.5\%/2.5\%$ over the strong baseline CLIP.
\end{itemize}
\section{Related Work}
\noindent\textbf{Unsupervised Domain Adaptation.} Unsupervised Domain Adaptation (UDA) adapts a model trained on a labeled source domain to an unlabeled target domain.
Quite a few UDA methods learn domain-invariant features via minimizing the discrepancy between domains~\cite{long2015learning, long2017deep, sun2016deep}. For example, Tzeng \textit{et~al.} \cite{tzeng2014deep} introduce an adaptation layer and a domain confusion loss to learn semantically meaningful and domain-invariant representations. DAN \cite{long2015learning} aligns source and target domains by minimizing the maximum mean discrepancy (MMD) on task-specific layers. Sun \textit{et~al.} \cite{sun2016deep} propose CORAL that aligns the second-order statistics of the source and target domain with a linear projection.

Inspired by generative adversarial networks (GANs) \cite{GAN_NIPS14},  another family of UDA methods apply adversarial learning to obtain domain-invariant representations \cite{ganin2015unsupervised, long2017conditional, JADA_MM19}. 
For example, DANN \cite{ganin2015unsupervised} and CDAN \cite{long2017conditional} introduce a domain discriminator to distinguish source samples from target ones, while the feature extractor tries to generate domain-invariant features in order to fool the domain discriminator. Differently, MCD \cite{saito2018maximum} plays the minimax game between a feature encoder and two classifiers, where two classifiers try to maximize their prediction discrepancy and the feature extractor aims to minimize that discrepancy.

Despite the success achieved by domain alignment, class discrimination also loses due to the distorted structure of semantic features~\cite{cai2019learning,tang2020unsupervised}. How to maintain class discriminability has also been considered by recent UDA works \cite{tang2020unsupervised, TPN_CVPR2019,li2020domain, ETD_CVPR20, BNM_CVPR2020, DWL_CVPR21}. To name a few, Li \textit{et al.} \cite{ETD_CVPR20} build attention-aware transport distance to learn discriminant features, along with an entropy-based regularization. Cui \textit{et al.} \cite{BNM_CVPR2020} propose to enforce the prediction discriminability and diversity via batch nuclear-norm maximization (BNM). However, these methods have to make trade-offs between aligning domains and preserving class discriminability. 

Compared with these methods, our method applies prompt learning to learn domain-specific visual concepts (\textit{i.e.}, the transparent background for ``product" domain) for each domain.



\noindent \textbf{Prompt Learning.} Prompt learning, which is first introduced by Petroni~\itshape{et~al.}\upshape~\cite{ DBLP:conf/emnlp/PetroniRRLBWM19}, has been widely studied in NLP during these years~\cite{DBLP:journals/corr/abs-2107-13586,DBLP:conf/emnlp/ShinRLWS20,DBLP:journals/corr/abs-2104-08691,DBLP:conf/acl/LiL20,DBLP:journals/tacl/JiangXAN20,DBLP:conf/emnlp/PetroniRRLBWM19}. Prompting means prepending instructions to the input and pre-training the language model so that the downstream tasks can be promoted. Petroni~\itshape{et~al.}\upshape~\cite{ DBLP:conf/emnlp/PetroniRRLBWM19} and P{\"{o}}rner~\itshape{et~al.}\upshape~\cite{ DBLP:journals/corr/abs-1911-03681} use manually defined prompts to improve the performance of language models. However, manually created prompts may be sub-optimal or even inappropriate, which might fail to provide accurate instruction. To obtain more accurate estimation of the knowledge contained in language models, several methods have been proposed to automatically explore optimal prompts~\cite{DBLP:journals/tacl/JiangXAN20, DBLP:conf/emnlp/ShinRLWS20,DBLP:conf/naacl/ZhongFC21}. More recently, prompts have been integrated into vision-language models to learn generic visual representations~\cite{radford2021learning,DBLP:conf/icml/JiaYXCPPLSLD21,DBLP:journals/corr/abs-2109-01134}. Among them, ALIGN~\cite{DBLP:conf/icml/JiaYXCPPLSLD21} and CLIP~\cite{radford2021learning} are most pioneering ones. CLIP~\cite{radford2021learning} learns state-of-the-art visual representations from natural language supervision by pre-training a vision language model on 400 million image-text pairs. Furthermore, Zhou~\itshape{et~al.}\upshape~\cite{ DBLP:journals/corr/abs-2109-01134} use continuous representations to model prompts so that the task-relevant prompts can be automatically learned, namely CoOp. However, CoOp only develops a domain-agnostic prompt for visual recognition tasks while our work proposes to learn both domain-agnostic and domain-specific prompts to deal with distribution shift in UDA.


\section{Method}

Given a set of labeled source images $\mathcal{D}_s = \{(\rvx_i^s, y_i^s)\}_{i=1}^{N_s}$ and a set of unlabeled target images $\mathcal{D}_u = \{(\rvx_i^u)\}_{i=1}^{N_u}$, we adopt a model trained from a source domain to a target domain.
Here, $N_s$ and $N_u$ denote the scale of source domain dataset $\gD_s$ and target domain dataset $\gD_u$ respectively. These two domains share the same $K$ categories.


\subsection{Preliminaries}
\label{sec:p}

\begin{figure*}
    \centering
    \includegraphics[width=0.90\linewidth]{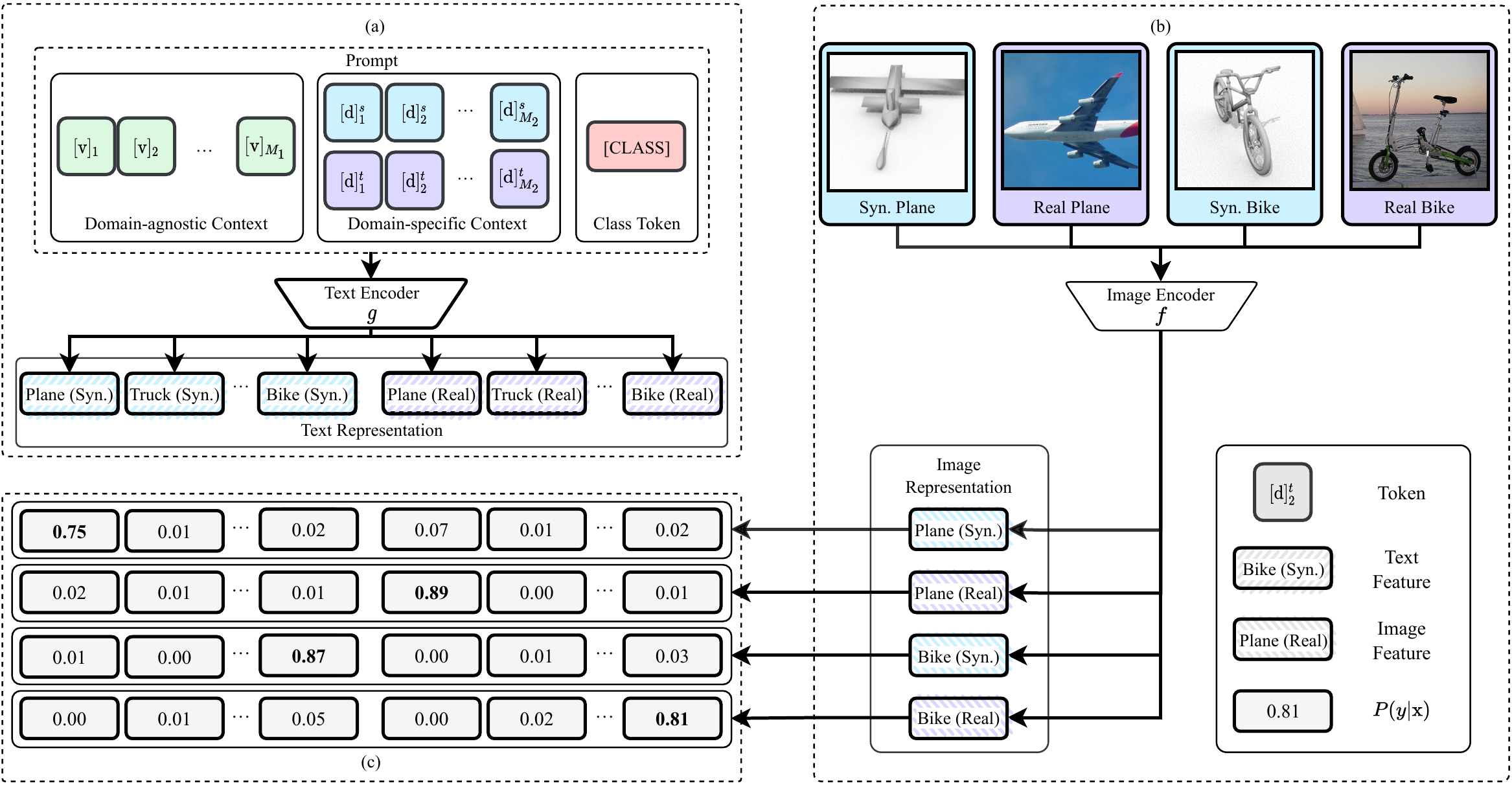}
    \caption{\textbf{Domain Adaptation via Prompt Learning (DAPL):} (a) DAPL trains the learnable context variables: domain-agnostic context variables and domain-specific context variables, and [CLASS] token which are combined and encoded by a text encoder. (b) An image encoder encodes images from different domains. (c) Next, cosine similarity between text and image features is computed and the positive pairs (with matched domain and class) are encouraged to align.  The classification probability are defined in \cref{eq:pro} and a cross-entropy loss is applied between the image feature and the ground truth class to train the networks. 
    }
    \label{fig:3}\vspace{-3mm}
\end{figure*}


We adopt CLIP~\cite{radford2021learning} as our backbone. Our model is comprised of an image encoder $f(\cdot)$ and a text encoder $g(\cdot)$. The image encoder can be a ResNet~\cite{he2016deep} or Vision Transformer (ViT)~\cite{dosovitskiy2020image}, and the text encoder is a Transformer~\cite{vaswani2017attention}. The image and text input can be directly transformed from high dimensional space into a low dimensional feature space by the encoders.

CLIP~\cite{radford2021learning} is trained with image-text pairs in a contrastive manner. Each input text describes a category in the format of ``a photo of a [CLASS]"~([\textrm{CLASS}] is the class token). A positive pair is an image $\rvx_i$ with its corresponding text $\rvt_i$ describing the category of $\rvx_i$. A negative pair is an image $\rvx_i$ with an irrelevant description $\rvt_j, j\neq i$ in the mini-batch. The training objective is to maximize the cosine similarity of positive pairs and minimize the cosine similarity of negative pairs. The contrastive learning objective aligns the image and text representation in the same feature space. 


With the aligned features, the model is capable of performing zero-shot inference. By forwarding $K$ category descriptions, an image $\rvx$ would belong to the category $\hat y_i$ with the largest similarity:
\begin{align}
    & P(\hat y=i|\rvx) = \frac{\exp (\langle g(\rvt_i), f(\rvx)\rangle /T)}{\sum_{k=1}^K \exp (\langle g(\rvt_k), f(\rvx)\rangle /T)} \label{eq:1},\\
    & \hat y_i = \arg\max_{k} P(\hat y_i = k),
\end{align}
where $T$ is a user-defined hyper-parameter (temperature) and $\langle \cdot, \cdot\rangle $ denotes the cosine similarity. 


The input text described above is a manually designed prompt comprised of a sequence of discrete tokens. The manually designed prompts are transformed into fixed vectors in the word embedding space. Since these vectors could be sub-optimal for the representation of categories, we could optimize the continuous embedding of the tokens. The continuous representation $\rvt_k$ allows for a more precise description of semantic features which are important to the context variable learning. 

Existing prompt learning methods adopt a domain-agnostic style that context is shared across all domains and all categories. It follows a unified style:
\begin{equation}
    \rvt_k = [\rvv]_1[\rvv]_2 \ldots [\rvv]_{M_1} [\mathrm{CLASS}]_k, \label{eq:unified}
\end{equation}
where $[\rvv]_{m_1}, m_1 \in  \{1, 2, \ldots, M_1\}$ is a vector with the same dimension as the word embedding, and $M_1$ is the number of context tokens applied in the prompt. 

\subsection{Domain Adaptation via Prompt Learning}\label{sec:dapl}


Since the domain-agnostic context alone cannot deal with the distribution shift between domains, we propose to use Domain-Specific Context~(DSC) to capture unique features of each domain. To be specific, our proposed prompt contains two counterparts, a domain-agnostic context and a domain-specific context. We use $[\rvd]_{m_2}^d, m_2 \in \{1, 2, \ldots, M_2\}$ to denote domain-specific tokens, which have the same dimension as word embeddings. The domain-specific context is shared among all categories but specially designed for each domain $[\rvd]_i^s \neq [\rvd]_j^u, i,j \in \{1, 2, \ldots, M_2\}$. The number of domain-specific tokens is denoted by $M_2$. Domain indicator denotes the source and target domains $ d\in \{s, u\}$. The overall prompt is defined in the following format:
\begin{equation}
    \rvt^d_k = [\rvv]_1[\rvv]_2 \ldots [\rvv]_{M_1} [\rvd]^d_1 [\rvd]^d_2 \ldots [\rvd]^d_{M_2} [\mathrm{CLASS}]_k.
    \label{eq:3}
\end{equation}

When $[\mathrm{CLASS}]$ token in the text feature space could not fully model the difference among each class, the domain-agnostic context could follow a class-specific style \cite{radford2021learning} denoted by class-specific context. Each class could be initialized with different tokens:
\begin{equation}
    \rvt^d_k = [\rvv]_1^k[\rvv]_2^k \ldots [\rvv]_{M_1}^k [\rvd]^d_1 [\rvd]^d_2 \ldots [\rvd]^d_{M_2} [\mathrm{CLASS}]_k. \label{eq:cl}
\end{equation}

The trainable class-specific context could learn a more fine-grained representation than only $[\mathrm{CLASS}]$ token~\cite{DBLP:journals/corr/abs-2109-01134}. Our main results are based on class-specific context and domain-specific context as \cref{eq:cl}. 

We have $2K$ categories since we apply different prompts $\rvt_k^s, \rvt^u_k$ for the source and the target domain respectively. Given a set of training samples $\{\rvx_i^s, y_i^s\}_{i=1}^{N_s}$ of the source domain, 
we could obtain the probability that a training sample belongs to the $k$-th category:
\begin{equation}
    P(\hat y_i^s=k|\rvx_i^s) = \frac{\exp (\langle g(\rvt_k^s), f(\rvx_i^s)\rangle /T)}{\sum_{d \in \{s,u\}}\sum_{j=1}^K \exp (\langle g(\rvt_j^d), f(\rvx_i^s)\rangle /T)}. \label{eq:pro}
\end{equation}

With the probability of the image $\rvx_i$ belonging to class $k$, we minimize the standard cross-entropy loss given ground truth label $y_i^s$. The loss is computed as follow: 
\begin{equation}\vspace{-1mm}
    \gL_s = - \frac{1}{N_s}  \sum_{i = 1}^{N_s} \log P(\hat y_i^s = y_i^s).
\end{equation}\vspace{-1mm}
\begin{figure}
    \centering
    \includegraphics[width=0.9\linewidth]{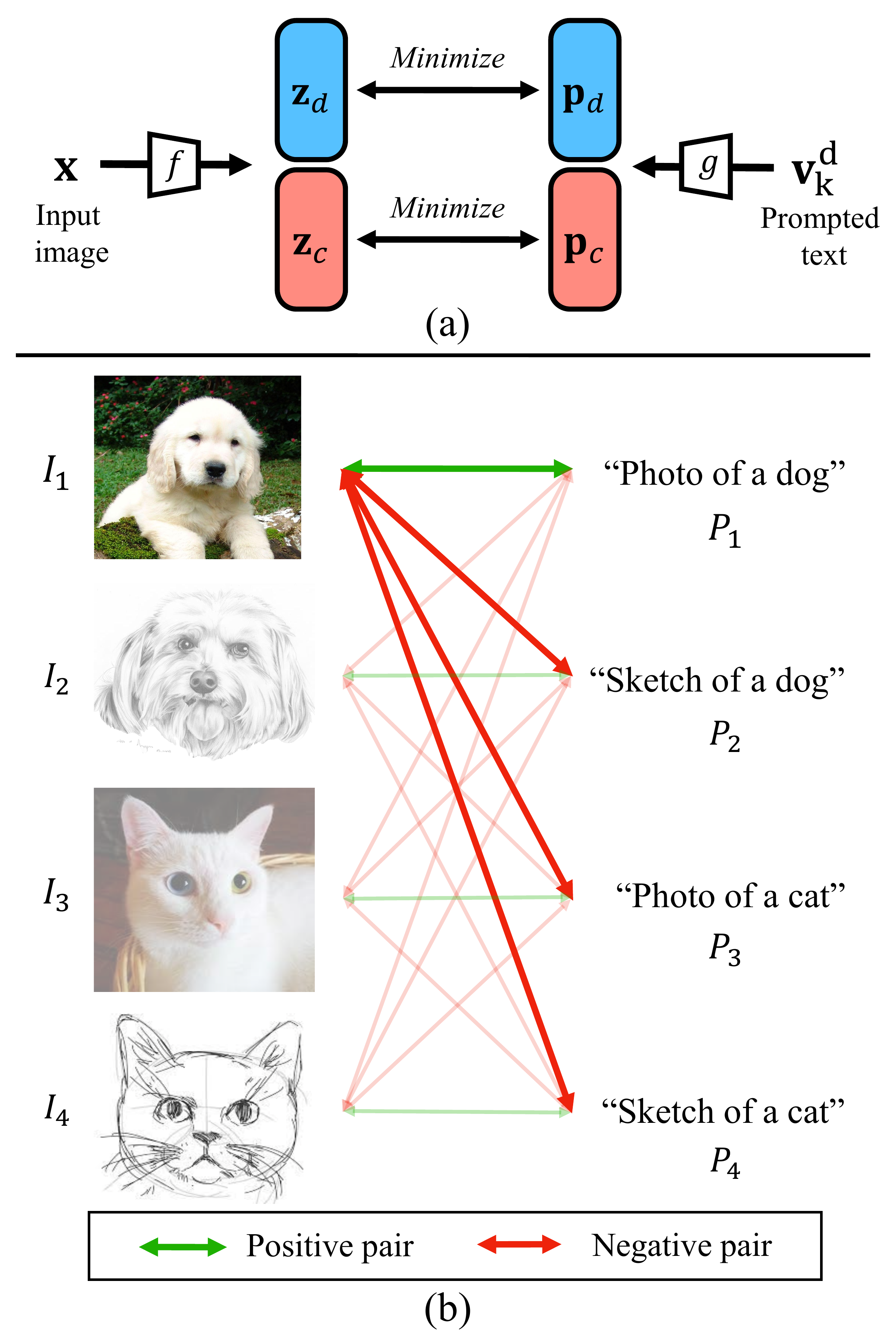}
    \vspace{-3mm}
    \caption{\textbf{Contrastive learning helps transfer learning.} (a) We assume that visual representation implicitly contains two parts: domain information ($\mathbf{z}_d$) and class information ($\mathbf{z}_c$). Similarly, the language feature contains two parts: domain information ($\mathbf{p}_d$) and class information ($\mathbf{p}_c$). By minimizing the distance between positive pairs (shown in green) and maximizing the distance between negative pairs (shown in red), we show that the domain information and class information can be disentangled. Such disentangled representations can be applied for transfer learning. See~\cref{sec:disentangle} for details. }
    \label{fig:contrastive_transfer}
    \vspace{-4mm}
\end{figure}
To further exploit the unlabeled data, We generate pseudo labels on the target domain. We choose from $K$ classes with maximum predicted probability as the pseudo label $y^u$ of the training data $\rvx^u$:
\begin{align}
    y^{u} = \arg\max_{k} P(\hat y^u = k| \rvx^u), \ k = \{1, 2, \ldots, K\}.
\end{align}

\vspace{-1mm}We only generate pseudo labels for unlabeled data whose maximum prediction probability is larger than a fixed threshold $\tau$ for the quality of pseudo labels. We make use of the zero-shot inference ability of CLIP to generate pseudo labels as described in \cref{sec:p}. We train the prompt of target domain $\rvt_k^u$ with these unlabeled images and their pseudo labels with the contrastive objective \cref{eq:pro}:
\vspace{-1mm}
\begin{equation}
\gL_{u}\!\! = \!-\frac{1}{N_u}\!\! \sum_{i = 1}^{N_u}\!\mathbb{I}\{P(\hat y^u_i\! = y^{u}_i| \rvx^u_i) \!\!\ge\!\! \tau \!\}\! \log\! P(\hat y^u_i\!\! =\! y^{u}_i| \rvx^u_i),\!\!\! \vspace{-1mm}
\end{equation}
where $\mathbb{I}\{\cdot\}$ is an indicator function. Overall, our proposed \textbf{Domain Adaptation via Prompt Learning~(DAPL)} method could be trained in an end-to-end manner with a total contrastive loss:
\vspace{-1mm}
\begin{equation}
    \gL = \gL_s(\gD^s) + \gL_{u}(\gD^u).\label{eq:totalloss} \vspace{-1mm}
\end{equation}


\vspace{-1mm}Existing domain adaptation methods train their classifier on the source domain to learn a conditional probability distribution $P(y|\rvx^s)$. By aligning the marginal distribution of $P(f(\rvx^s))$ and $P(f(\rvx^u))$ they could directly make use of the conditional probability for inference on the target domain. When the conditional probability distribution varies $P(y|\rvx^s) \neq P(y|\rvx^u)$, these methods could suffer the risk of performance drop~\cite{wang2020transfer}. Our method does not align marginal distributions but learns two conditional probability distributions $P(y|\rvx^s)$ and $P(y|\rvx^u)$ by learning two sets of prompts $\rvt_k^s, \rvt^u_k, k \in \{1, 2, \ldots, K\}$. Hence, our method could deal with both conditional distribution shift and marginal distribution shift. The overview of DAPL is shown in \cref{fig:3}. 
\vspace{-2mm}

\subsection{Disentanglement by Contrastive Learning}\label{sec:disentangle}

We adopt a contrastive loss $\gL$ as the optimization objective. 
Here, we provide an intuitive explanation for why this objective achieves the desired goal: the visual encoder and text encoder each encodes the input into two disentangled latent representations, separating domain information from the intrinsic class information. Only when both the class and the domain information are aligned, the distance between the textual feature and the image feature is minimized. By minimizing the distance between such positive pairs (maximizing the similarity), the probability of the correct label is maximized (see \cref{eq:pro}).

First, we assume that the visual representation $f(\rvx_i^d)$ contains two parts: domain information of domain $d$ and the intrinsic class information of class $c$ (\cref{fig:contrastive_transfer} (a), $\rvz_d$ and $\rvz_c$).  Similarly, the language embedding $g(\rvt_k^d)$ contains the same two parts: domain information of domain $d$ and the class information of class $c$ (\cref{fig:contrastive_transfer} (a),  $\rvp_d$ and $\rvp_c$). Next, we show that such domain information and class information can be disentangled by optimizing the contrastive objective. 

Figure~\ref{fig:contrastive_transfer} (b) provides an illustrative example. In this example, there are four image-text pairs with two classes (\textit{cat}, \textit{dog}) and two domains (\textit{photo}, \textit{sketch}). 
Take the image $I_1$, prompts $P_{1}$ and $P_{2}$ as an example. The image can form a positive pair with prompt $P_1$ and a negative pair with prompt $P_{2}$. By optimizing the contrastive objective, the distance between image feature $f(I_1)$ and the sentence embedding of $g(P_1)$ is minimized, whereas the distance between image feature $f(I_1)$ and the sentence embedding of $g(P_2)$ is maximized. We claim that this forces the class information of \textit{dog} disentangled from the domain representation of \textit{photo} or \textit{sketch}. Suppose on the contrary that the domain information and the class information are still \textit{entangled} in the representation,  \textit{i.e.} the domain representation ($\rvp_d^1$ and $\rvp_d^2$) contains the class information of \textit{dog}. In this case, $I_1$ and $P_2$ still matches and the distance between $f(I_1)$ and $g(P_2)$ could be further maximized by removing this class information. In other words, we reduce class information in domain representation by optimizing the contrastive loss. Similarly, taking $(I_1, P_3)$ as negative pair, we remove domain information from class representation - otherwise $f(I_1)$ still matches $g(P_3)$ because of the \textit{entangled} domain information of \textit{photo} in class representation. Combining these two negative pairs, the domain representation and the intrinsic class information can be forced to disentangle with each other by minimizing the contrastive objective.

\begin{table*}[!ht]
    \centering
    \setlength{\abovecaptionskip}{0.cm}
    \setlength{\belowcaptionskip}{0.cm}
    \caption{Accuracy (\%) on Office-Home\cite{Office-Home} for unsupervised domain adaptation (ResNet-50\cite{he2016deep}). The best accuracy is indicated in bold.}
    \setlength{\tabcolsep}{0.28mm}
    {
    \begin{tabular}{c|cccccccccccc|c}
        Method & Ar→Cl & Ar→Pr & Ar→Rw & Cl→Ar & Cl→Pr & Cl→Rw & Pr→Ar & Pr→Cl & Pr→Rw & Rw→Ar & Rw→Cl & Rw→Pr & Avg  \\ \shline
        ResNet-50\cite{he2016deep} & 34.9 & 50.0 & 58.0 & 37.4 & 41.9 & 46.2 & 38.5 & 31.2 & 60.4 & 53.9 & 41.2 & 59.9 & 46.1  \\ \hline
        DANN \cite{ganin2015unsupervised} & 45.6 & 59.3 & 70.1 & 47.0 & 58.5 & 60.9 & 46.1 & 43.7 & 68.5 & 63.2 & 51.8 & 76.8 & 57.6  \\ \hline
        JAN \cite{long2017deep} & 45.9 & 61.2 & 68.9 & 50.4 & 59.7 & 61.0 & 45.8 & 43.4 & 70.3 & 63.9 & 52.4 & 76.8 & 58.3 \\
        \hline
        CDAN+E \cite{long2017conditional} & 50.7 & 70.6 & 76.0 & 57.6 & 70.0 & 70.0 & 57.4 & 50.9 & 77.3 & 70.9 & 56.7 & 81.6 & 65.8 \\ \hline
        BSP+CDAN \cite{BSP_ICML2019} & 52.0 & 68.6 & 76.1 & 58.0 & 70.3 & 70.2 & 58.6 & 50.2 & 77.6 & 72.2 & 59.3 & 81.9 & 66.3 \\ \hline
        SymNets \cite{zhang2019domain} & 47.7 & 72.9 & 78.5 & 64.2 & 71.3 & 74.2 & 63.6 & 47.6 & 79.4 & 73.8 & 50.8 & 82.6 & 67.2  \\ \hline
        ETD \cite{ETD_CVPR20} & 51.3 & 71.9 & \textbf{85.7} & 57.6 & 69.2 & 73.7 & 57.8 & 51.2 & 79.3 & 70.2 & 57.5 & 82.1 & 67.3 \\
        \hline
        BNM \cite{BNM_CVPR2020} & 52.3 & 73.9 & 80.0 & 63.3 & 72.9 & 74.9 & 61.7 & 49.5 & 79.7 & 70.5 & 53.6 & 82.2 & 67.9 \\ \hline
        
        GSDA \cite{hu2020unsupervised} & \textbf{61.3} & 76.1 & 79.4 & 65.4 & 73.3 & 74.3 & 65.0 & 53.2 & 80.0 & 72.2 & \textbf{60.6} & 83.1 & 70.3  \\ \hline
        GVB-GD \cite{cui2020gradually} & 57.0 & 74.7 & 79.8 & 64.6 & 74.1 & 74.6 & 65.2 & \textbf{55.1} & 81.0 & 74.6 & 59.7 & 84.3 & 70.4  \\ \hline
        RSDA-MSTN \cite{gu2020spherical} & 53.2 & 77.7 & 81.3 & 66.4 & 74.0 & 76.5 & 67.9 & 53.0 & 82.0 & 75.8 & 57.8 & 85.4 & 70.9  \\ \hline
        SPL \cite{SPL_AAAI20} & 54.5 & 77.8 & 81.9 & 65.1 & 78.0 & 81.1 & 66.0 & 53.1 & 82.8 & 69.9 & 55.3 & \textbf{86.0} & 71.0 \\
        \hline
        SRDC \cite{tang2020unsupervised} & 52.3 & 76.3 & 81.0 & 69.5 & 76.2 & 78.0 & 68.7 & 53.8 & 81.7 & \textbf{76.3} & 57.1 & 85.0 & 71.3  \\ \bottomrule
        CLIP \cite{radford2021learning} & 51.6 & 81.9 & 82.6 & 71.9 & 81.9 & 82.6 & 71.9 & 51.6 & 82.6 & 71.9 & 51.6 & 81.9 & 72.0  \\ \hline
        \textbf{DAPL} & 54.1 & \textbf{84.3} & 84.8 & \textbf{74.4} & \textbf{83.7} & \textbf{85.0} & \textbf{74.5} & 54.6 & \textbf{84.8} & 75.2 & 54.7 & 83.8 & \textbf{74.5}
    \end{tabular}}
    \label{tab:officehome}
\end{table*}



\section{Experimental Results}

We conduct extensive experiments on UDA benchmarks to verify the validity of our proposed method. We next present the datasets used in our experiments, comparisons with baseline methods, ablation studies of our method and visualization of results. 

\subsection{Datasets and Experimental Settings}\label{sec:data}


\noindent \textbf{Office-Home} \cite{Office-Home} is a large-scale benchmark for visual cross-domain recognition. It collects a total of 15,500 images from four distinct domains: Art (\textit{Ar}), Clip Art (\textit{Cl}), Product (\textit{Pr}), and Real World (\textit{Rw}). Besides, each domain contains the objects of 65 categories in the office and home environments. To evaluate our method, we conduct 12 UDA tasks, \ie., Ar $\rightarrow$ Cl, ..., Rw $\rightarrow$ Pr. 

\noindent \textbf{VisDA-2017} \cite{VisDA-2017} is a more challenging dataset for synthetic-to-real domain adaptation with 12 categories. It contains 152,397 synthetic images, generated by rendering the 3D models with different angles and light conditions, and 55,388 real-world images, collected from MSCOCO \cite{MSCOCO}. Following \cite{long2017conditional} and \cite{saito2018maximum}, we use the synthetic images as source domain and real-world images as target domain.

\noindent \textbf{Implementation details.} For Office-Home, we use pre-trained CLIP model and adopt ResNet-50 \cite{He2016DeepRL} as its image encoder. We fix the parameters in the encoders and the prompt is trained with the mini-batch SGD optimizer for 200 epochs, where the batch size is set to be 32. The initial learning rate is set to 0.003 and decayed with a cosine annealing rule \cite{Loshchilov2017SGDRSG}. For VisDA-2017\cite{VisDA-2017}, the results are obtained by leveraging the pre-trained CLIP model with ResNet-101 \cite{He2016DeepRL} as the image encoder. The parameters of the image and text encoders are fixed and we train the prompt for 25 epochs using the mini-batch SGD optimizer with a batch of 32. The learning rate is set to 0.003 initially and decayed with a cosine annealing rule. As for the hyper-parameters, the length of context tokens $M_1$ and domain-specific tokens $M_2$ are both set to 16. Other choices of token numbers are discussed in \cref{sec:ablation}. Our context vectors are randomly initialized using a zero-mean Gaussian distribution with a standard deviation of 0.02. The pseudo labeling threshold $\tau$ is set to 0.6 for Office-Home and 0.5 for VisDA-2017\cite{VisDA-2017}. Further discussion about the value of $\tau$ is shown in \cref{sec:ablation}. 

\begin{table*}[h]
    \centering
    \setlength{\abovecaptionskip}{0.cm}
    \setlength{\belowcaptionskip}{0.cm}
     \caption{Accuracy (\%) on VisDA-2017\cite{VisDA-2017} for unsupervised domain adaptation (ResNet-101\cite{he2016deep}). The best accuracy is indicated in bold.}
    \setlength{\tabcolsep}{1.86mm}
    {
    \begin{tabular}{c|cccccccccccc|c}
        Method & plane & bicycle & bus & car & horse & knife & mcycl & person & plant & sktbrd & train & truck & Avg \\ \shline
        ResNet-101 \cite{he2016deep} & 55.1 & 53.3 & 61.9 & 59.1 & 80.6 & 17.9 & 79.7 & 31.2 & 81.0 & 26.5 & 73.5 & 8.5 & 52.4 \\ \hline
        DANN \cite{ganin2015unsupervised} & 81.9 & 77.7 & 82.8 & 44.3 & 81.2 & 29.5 & 65.1 & 28.6 & 51.9 & 54.6 & 82.8 & 7.8 & 57.4 \\ \hline
        JAN \cite{long2017deep} & 75.7 & 18.7 & 82.3 & \textbf{86.3} & 70.2 & 56.9 & 80.5 & 53.8 & 92.5 & 32.2 & 84.5 & 54.5 & 65.7  \\ \hline
        MCD \cite{saito2018maximum} & 87.0 & 60.9 & 83.7 & 64.0 &  88.9 & 79.6 & 84.7 & 76.9 &  88.6 & 40.3 & 83.0 & 25.8 & 71.9 \\ 
        \hline
        CDAN+E \cite{long2017conditional} & 85.2 &  66.9 & 83.0 & 50.8 & 84.2 & 74.9 & 88.1 & 74.5 & 83.4 & 76.0 & 81.9 & 38.0 & 73.9 \\
        \hline
        BSP+CDAN \cite{BSP_ICML2019} & 92.4 & 61.0 & 81.0 & 57.5 & 89.0 & 80.6 & 90.1 & 77.0 & 84.2 & 77.9 & 82.1 & 38.4 & 75.9 \\ \hline
        SWD \cite{SWD_CVPR19} & 90.8 & 82.5 & 81.7 & 70.5 & 91.7 & 69.5 & 86.3 & 77.5 & 87.4 & 63.6 & 85.6 & 29.2 & 76.4  \\ \hline
        DWL \cite{DWL_CVPR21} & 90.7 & 80.2 & 86.1 & 67.6 & 92.4 & 81.5 & 86.8 & 78.0 & 90.6 & 57.1 & 85.6 & 28.7 & 77.1 \\ \hline
        MODEL \cite{li2020model} & 94.8 & 73.4 & 68.8 & 74.8 & 93.1 & \textbf{95.4} & 88.6 & \textbf{84.7} & 89.1 & 84.7 & 83.5 & 48.1 & 81.6  \\ \hline
        CGDM \cite{CGDM_CVPR21} & 93.4 & 82.7 & 73.2 & 68.4 & 92.9 & 94.5 & 88.7 & 82.1 &  93.4 & 82.5 & 86.8 & 49.2 & 82.3 \\
        \hline
        STAR \cite{lu2020stochastic} & 95.0 & \textbf{84.0} & 84.6 & 73.0 & 91.6 & 91.8 & 85.9 & 78.4 & \textbf{94.4} & 84.7 & 87.0 & 42.2 & 82.7  \\ \bottomrule
        CLIP \cite{radford2021learning} & \textbf{98.2} & 83.9 & \textbf{90.5} & 73.5 & 97.2 & 84.0 & \textbf{95.3} & 65.7 & 79.4 & \textbf{89.9} & 91.8 & \textbf{63.3} & 84.4 \\ \hline
        \textbf{DAPL} & 97.8 & 83.1 & 88.8 & 77.9 & \textbf{97.4} & 91.5 & 94.2 & 79.7 & 88.6 & 89.3 & \textbf{92.5} & 62.0 & \textbf{86.9} 
    \end{tabular}}
    \label{tab:visda}
\end{table*}

\subsection{Comparison with State-of-the-Art DA Methods}
\subsubsection{Quantitative Evaluation}
\noindent \textbf{Results on Office-Home} are shown in \cref{tab:officehome}, where our method obviously outperforms all other baselines w.r.t the average accuracy of 12 tasks. Note that there exists a large performance gap between the feature alignment-based methods (e.g., DANN \cite{ganin2015unsupervised} and CDAN+E \cite{long2017conditional}) and SRDC \cite{tang2020unsupervised}. The possible reason may be that excessive feature alignment would hamper the discrimination of target data. While such potential risk will not happen in our method, since we do not force feature alignment across domains. Particularly, our method further surpasses the state-of-the-art method SRDC \cite{tang2020unsupervised}) by a large margin of 3.2\% in terms of the average accuracy. 
We owe the performance improvement to the more suitable visual concepts for the target domain that are generated from our learned prompts. And the superior performance of our method shows that simple prompt learning is effective for UDA problems.

\noindent \textbf{Results on VisDA-2017\cite{VisDA-2017}} are presented in \cref{tab:visda}. It can be observed that our method achieves the highest average accuracy of $86.9\%$ over the 12 classes, outperforming the state-of-the-art method STAR \cite{lu2020stochastic} by a large margin of $4.2\%$. Note that CLIP in \cref{tab:visda} means zero-shot CLIP which adopts ``a photo of a [CLASS]" as the hand-crafted prompt. Even the hand-crafted prompt method already has an impressive performance, our DAPL still achieves a $2.5\%$ absolute improvement over it. The reason why the accuracy of truck is significantly boosted may be that the concept of ``truck'' is more discriminative in the language model.
Furthermore, with the help of prompt learning, DAPL outperforms CLIP by $7.5\%, 14\%, 9.2\%$ on ``knife", ``person" and ``plant". In general, despite the simplicity of ours method, the encouraging results validate the efficacy of our prompt learning method.

\subsubsection{Training Time Analysis}
We train all the models with 1 NVIDIA RTX 2080 Ti GPU. Our method is much more efficient than other methods. For example, DAPL, MCD\cite{saito2018maximum} and DANN\cite{ganin2015unsupervised} take 5.3h, 13.4h, 38.3h to train on VisDA-2017, respectively. Because we only fine-tune the prompt with very few parameters, it is much easier and faster to optimize the model.

\begin{table}[h]
    \centering
    \tablestyle{10pt}{1.2}
    \caption{\textbf{Ablation: the effectiveness of domain-specific context (DSC).} Domain-specific context is crucial for achieving good performance. The numbers show classification accuracy~($\%$) on VisDA-2017\cite{VisDA-2017} dataset. Higher values are better. The numbers in brackets show absolute improvement from baseline. }
    {
    \begin{tabular}{cc|c}
         
        Domain-agnostic & Domain-specific & Cls. Acc. \\
        \shline
         Manual & \xmark & 84.4 \\ \hline
         Unified & \xmark  & 85.5 {\color{OliveGreen}(+1.1)} \\ 
         Class-specific & \xmark & 86.2 {\color{OliveGreen}(+1.8)} \\ \hline
         Unified & \cmark & \textbf{86.9} {\color{OliveGreen}(+2.5)} \\ 
         Class-specific &  \cmark & \textbf{86.9} {\color{OliveGreen}(+2.5)}
    \end{tabular}}
    \label{tab:dsc}
    \vspace{-1em}
\end{table}
\begin{figure}
    \centering
    \includegraphics[width=\linewidth]{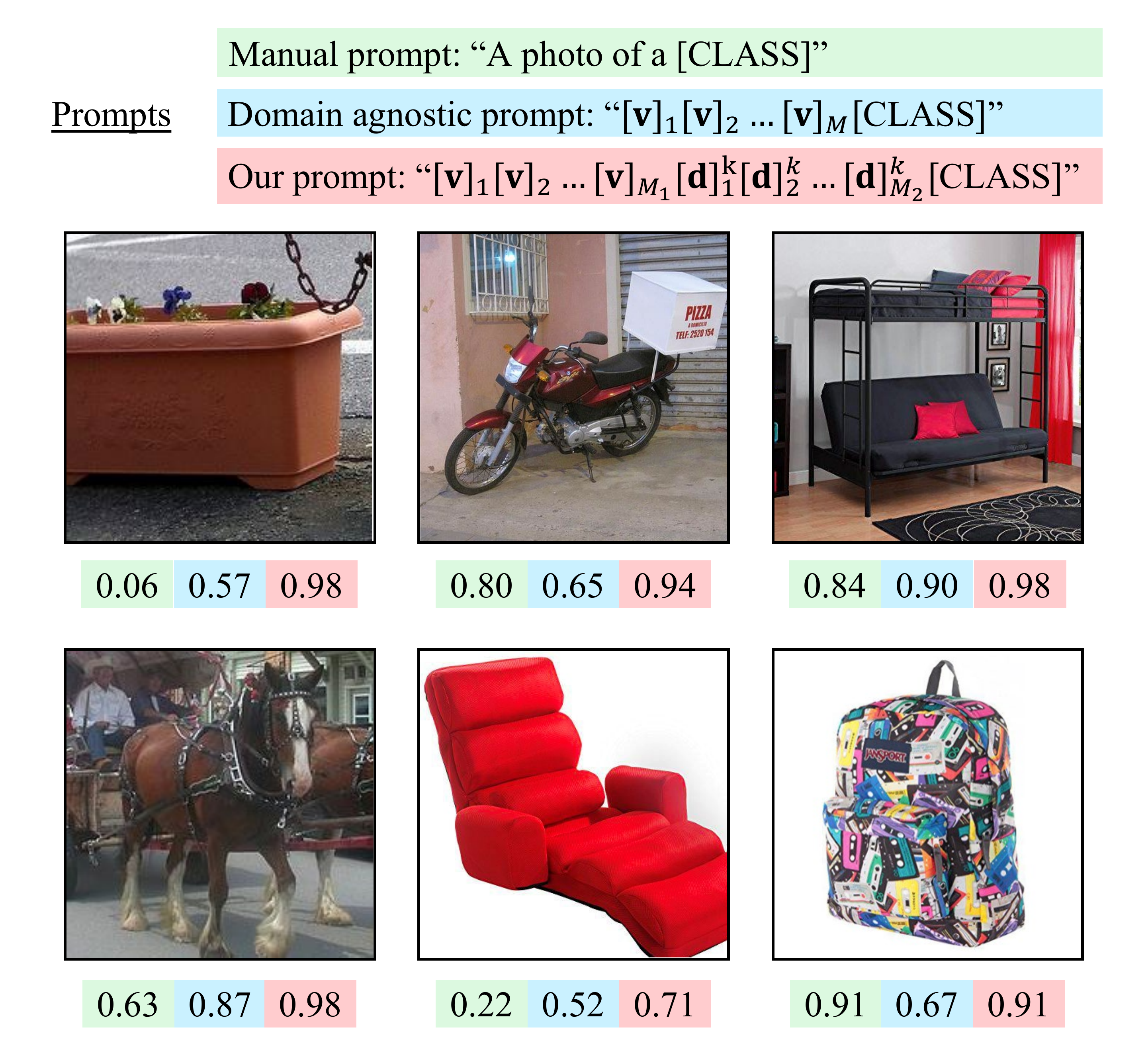}
    \caption{\textbf{Prediction confidence from VisDA-2017 (\emph{top}) and Office-Home dataset (\emph{bottom})}. Confidence \emph{of the ground-truth class} predicted using different prompting methods.
    Blue: manually designed prompt. 
    Green: domain-agnostic prompt.  
    Pink: our proposed method. Predictions given by our method show the highest confidence.}
    \label{fig:vis}\vspace{0mm}
\end{figure}

\subsection{Ablation Study} \label{sec:ablation}

To give a more detailed analysis of our method, we conduct several ablation studies on VisDA-2017\cite{VisDA-2017}. All of the variant models are trained with the same training hyper-parameters as described in \cref{sec:data}. 

\noindent \textbf{Ablation: domain-specific context.} To prove the effectiveness and necessity of domain-specific context, we compare the performances of these following prompt settings on VisDA-2017\cite{VisDA-2017} dataset: (1) the manually designed prompt ``a photo of [CLASS]" as the baseline; (2) the domain-agnostic prompt in the form of unified context~(as shown in \cref{eq:unified}); (3) the domain-agnostic prompt in the form of class-specific context; (4) the domain-agnostic prompt in the form of unified context with domain-specific context~(as shown in \cref{eq:3}); and (5) the domain-agnostic prompt in the form of class-specific context with domain-specific context~(as shown in \cref{eq:cl}).

The results of the above experiments are listed in \cref{tab:dsc}. Even the manually design prompt is a strong baseline, our proposed DAPL (4) and (5) achieves $2.5\%$ absolute improvement than the hand-crafted baseline (1). By comparing (2) with (3),  we can observe that learning prompt with class-specific context can have a better performance than with unified context when domain-specific context is not used. Because the differences between classes can be better modeled by the class-specific context. Combining domain-specific context with the unified context (\ie, (4)) can further bring $1.4\%$ performance improvement to (2). Besides, consistent performance improvement is also attained from (3) to (5). These improvements over the domain-agnostic context alone demonstrate the necessity of domain-specific context, which helps to capture the unique underlying domain information. Finally, by comparing (4) with (5), we know that tuning class-specific context with domain-specific context does not still yield improvement like (2) over (1). This is because distribution shift is the predominant factor in UDA, and modeling fine-grained discrepancy between classes may not further improve the performance. Thus, we choose the combination of unified context and domain-specific context in the paper. 

\noindent \textbf{Ablation: context token length.}
We conduct experiments in \cref{tab:length of context tokens} to explore the influence of context token length. The lengths of domain-agnostic and domain-specific context tokens are denoted by $M_1$ and $M_2$, respectively. From the results, we can see that the performance is a little lower when $M_1 < M_2$. Overall, the token length has little effect on the performance of our method. This implies the continuous representation could be learned with a small number of tokens. 

\begin{table}[h]
    \centering
    \tablestyle{4pt}{1.2}
    \caption{\textbf{Ablation: context token length.} The accuracy~($\%$) of different length combinations on VisDA-2017\cite{VisDA-2017} dataset (with ResNet-101 as image encoder). The values shown are ($M_1$, $M_2$), \ie., context length of domain-agnostic prompt and domain-specific prompt. The best performance is denoted in bold.}
    {
    \begin{tabular}{c|ccccc}
        \makecell[c]{Content token length} & (4, 28) & (8, 24) & (28, 4) & (16, 16) & (24, 8) \\ \shline
        Cls. Acc. & 86.6
         & 86.8
         & \textbf{86.9}
         & \textbf{86.9}
         & \textbf{86.9} 
    \end{tabular}}
    \label{tab:length of context tokens}
\end{table}

\noindent \textbf{Ablation: pseudo label threshold.}
In \cref{tab:pseudo label threshold}, we present the sensitivity of our method to the hyper-parameter $\tau$ by ranging it from 0.4 to 0.7. It seems that our method is not sensitive to $\tau$ because of the trade-off between quality and quantity of pseudo labels. For example, when $\tau$ is set to 0.7, the model is trained with fewer but more confident pseudo labels and the quality of pseudo labels may make up the performance drop brought by the reduced quantity.

\begin{table}[h]
    \centering
    \tablestyle{13pt}{1.2}
    \caption{\textbf{Ablation: pseudo label threshold.} The accuracy~($\%$) of different threshold $\tau$ on VisDA-2017\cite{VisDA-2017} dataset (with ResNet-101 image encoder). The best performance is denoted in bold.}
    {
    \begin{tabular}{c|cccc}
        Threshold $\tau$ & 0.4 & 0.5 & 0.6 & 0.7 \\ \shline
        Cls. Acc.  & \textbf{86.9} & \textbf{86.9} & 86.7 & 86.6 
    \end{tabular}}
    \label{tab:pseudo label threshold}\vspace{-3mm}
\end{table}




\subsection{Visualization}

In \cref{fig:vis}, we compare the prediction confidence of the ground truth category on the target domain when using three different prompts: (a) a hand-crafted prompt; (b) the prompt with only domain-agnostic context; and (c) the prompt with domain-agnostic context and domain-specific context. 


For the third example of the top row, the plant only takes up a small area of the image. Hence, the prompt ``a photo of a plant'' is inappropriate for the image, while ``a photo of a plant with a pot'' might be a better match. Therefore, the hand-crafted prompt performs poorly on this example. In contrast, the learnable prompt yields a more confident prediction than the manually designed prompt. For the last image of the bottom row, it is a good match for the prompt ``a photo of a backpack''. The learnable domain-agnostic context performs worse than the manually designed prompt. By learning domain information of ``product'', the domain-specific context enables the model with more confidence to predict the image as a backpack. Overall, these comparison results with different prompts validate that learnable domain-agnostic and domain-specific contexts improve the performance of our model when combined.

\section{Conclusion}

In this paper, we introduce a novel prompt learning method for unsupervised domain adaptation, which is free of aligning features between domains as conventional methods do~\cite{long2015learning}. Instead, we design domain-specific context for each domain to advocate learning distinct domain representations of the source and the target domain. By making use of the prompt learning, We build a bridge between multi-modality methods and domain adaptation methods. Extensive results have demonstrated the advantage of our method. Prompt learning methods can be extended to other visual tasks in unsupervised domain adaptation in the future, \eg, semantic segmentation.

\section*{Acknowledgements}

This work is supported in part by the National Science and Technology Major Project of the Ministry of Science and Technology of China under Grants 2018AAA0100701, the NSFC under Grant 62022048, the Guoqiang Institute of Tsinghua University.

{\small
\bibliographystyle{ieee_fullname}
\bibliography{egbib}
}


\end{document}